\documentclass[letterpaper, 10 pt, conference]{ieeeconf}  

\IEEEoverridecommandlockouts                              

\usepackage{mfirstuc}
\usepackage{graphics} 
\usepackage{epsfig} 
\usepackage{mathptmx} 
\usepackage{times} 
\usepackage{amsmath} 
\usepackage{amssymb}  
\usepackage{subcaption}
\usepackage[font={small}]{caption}
\usepackage{siunitx}
\usepackage{verbatim}
\usepackage[noadjust]{cite}
\usepackage{siunitx}
\usepackage{booktabs}
\usepackage{url}
\usepackage{listings}
 \usepackage{ulem} 
\usepackage{subcaption}
\usepackage{multirow}
\renewcommand{\baselinestretch}{0.929}
\usepackage{tikz}
\title{\LARGE \bf VIBES: Vibro-Inertial Bionic Enhancement System\\in a Prosthetic Socket\\ }

\author{Alessia Silvia Ivani\authorrefmark{4}$^{1,2,3}$, Federica Barontini$^{1,2}$, Manuel G. Catalano$^{1}$,  Giorgio Grioli$^{1,2}$,\\  Matteo Bianchi $^{2,3}$ and Antonio Bicchi$^{1,2,3}$   
\thanks{*This work was supported by the European Research Council Synergy Grant Natural BionicS (NBS) project (Grant Agreement No. 810346), by the Italian Ministry of Education and Research (MIUR)  in the framework of the FoReLab project and Crosslab project (Departments of Excellence); by PNRR, M4 C2 I1.5 Ecosistema dell'Innovazione "Tuscany Health Ecosystem (THE)" - Ecosistema dell’innovazione sulle scienze e le tecnologie della vita in Toscana (CUP I53C22000780001) - Spoke 9, and the ERC Proof of Concept Wearable Integrated Soft Haptic Display for Prosthetics (WISH) project (Grant No. 101069179).}
\thanks{$^{1}$with  Soft Robotics for Human Cooperation and Rehabilitation, Istituto Italiano di Tecnologia, Genova 16163, Italy.}{}
\thanks{$^{2}$with Centro di ricerca E. Piaggio, University of Pisa, Pisa 56122, Italy.}
\thanks{$^{3}$ with Department of Information Engineering, University of Pisa, Pisa 56122, Italy.}
 \thanks{\authorrefmark{4} Corresponding author \tt\small alessia.ivani@iit.it}
}
\newcommand\copyrighttext{%
  \footnotesize \textcopyright\ \the\year{} IEEE. Personal use of this material is permitted.  Permission from IEEE must be obtained for all other uses, in any current or future media, including reprinting/republishing this material for advertising or promotional purposes, creating new collective works, for resale or redistribution to servers or lists, or reuse of any copyrighted component of this work in other works.}

\newcommand\copyrightnotice{%
\begin{tikzpicture}[remember picture,overlay]
\node[anchor=south,yshift=10pt] at (current page.south) {\fbox{\parbox{\dimexpr\textwidth-\fboxsep-\fboxrule\relax}{\copyrighttext}}};
\end{tikzpicture}%
}
\begin{document}
\maketitle
\thispagestyle{empty}
\pagestyle{empty}

\begin{abstract}
The use of vibrotactile feedback is of growing interest in the field of prosthetics, but few devices fully integrate this technology in the prosthesis to transmit high-frequency contact information (such as surface roughness and first contact) arising from the interaction of the prosthetic device with external items.
This study describes a wearable vibrotactile system for high-frequency tactile information embedded in the prosthetic socket.
The device consists of two compact planar vibrotactile actuators in direct contact with the user's skin to transmit tactile cues. 
These stimuli are directly related to the acceleration profiles recorded with two IMUS placed on the distal phalanx of a soft under-actuated robotic prosthesis (SoftHand Pro).
We characterized the system from a psychophysical point of view with fifteen able-bodied participants by computing participants' Just Noticeable Difference (JND) related to the discrimination of vibrotactile cues delivered on the index finger, which are associated with the exploration of different sandpapers.
Moreover, we performed a pilot experiment with one SoftHand Pro prosthesis user by designing a task, i.e. Active Texture Identification, to investigate if our feedback could enhance users' roughness discrimination.
Results indicate that the device can effectively convey contact and texture cues, which users can readily detect and distinguish.
\end{abstract}
\copyrightnotice
\section{Introduction}
Sensory feedback is widely recognized as a priority for prosthetic users \cite{cordella2016literature,Consumerdesign}, given its fundamental role in our cognitive processes, emotional state, and behaviour.
Extensive research efforts have been directed towards methods to reproduce haptic feedback after the loss or absence of a hand to relieve prosthetic users from the need for constant visual attention \cite{motamedi2016use}.
However, sensorized prostheses with feedback systems are currently only available through a few commercial options, as in \cite{akhtar2020touch}. 
Furthermore, developing a non-invasive sensorized prosthesis demands careful consideration of which haptic information transmit to the user and how, keeping the entire system intuitive, wearable, and integrated.

\begin{figure}
      \centering
      \includegraphics[width=0.9\columnwidth]{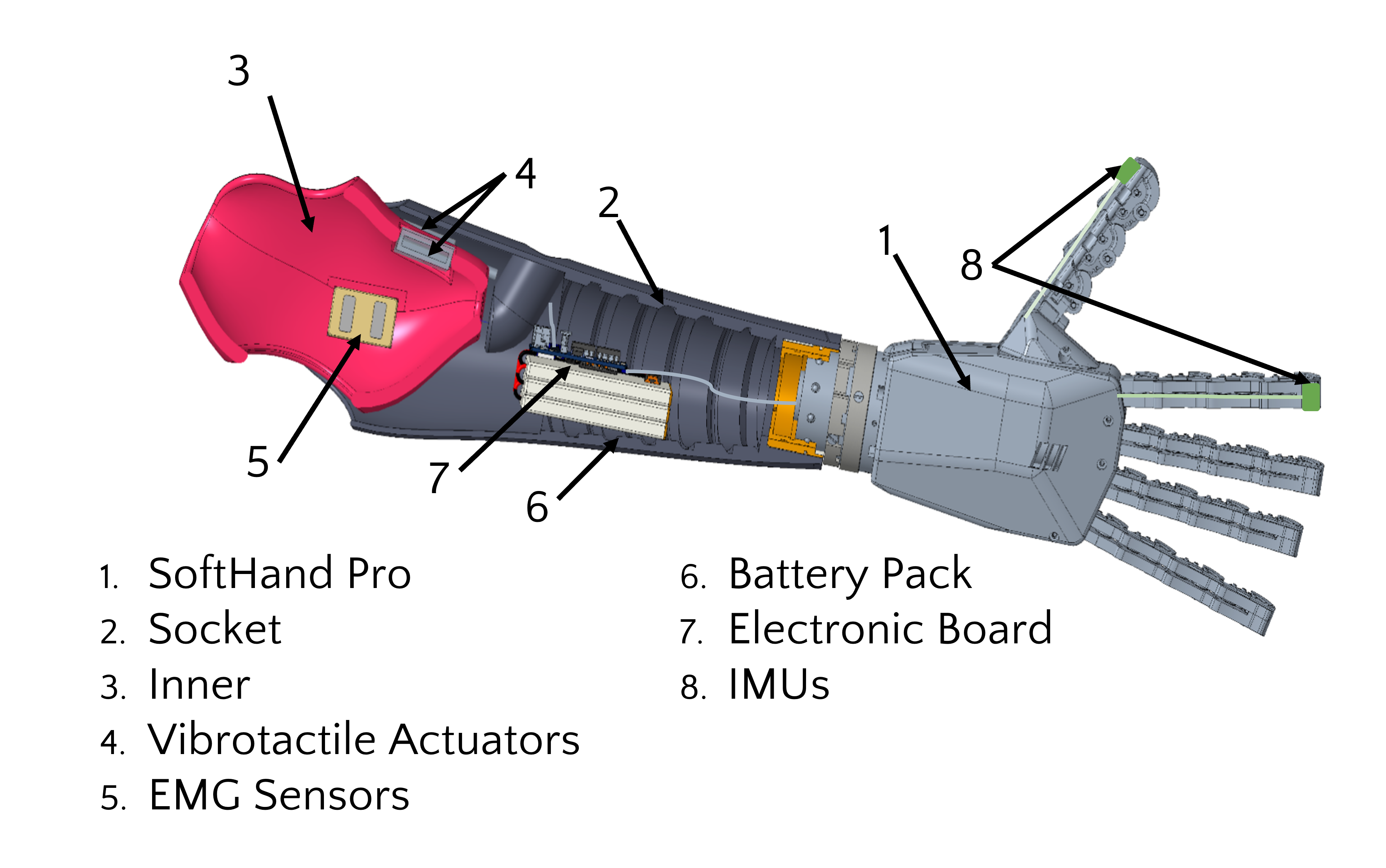}
      \caption{Overview of the main components of the VIBES: Vibro-Inertial Bionic Enhancement System.}
      \label{fig:system}
      \vspace{-0.7cm}
\end{figure} 
Non-invasive feedback can be conveyed through various methods, with the most prevalent ones being electrotactile, mechano-tactile (force stimuli), and vibrotactile stimulation.
Electrotactile stimulation is a non-invasive feedback solution that involves the application of electrical currents to the skin, thereby activating tactile nerve fibers~\cite{bensmaia2020restoration}.
The electrotactile electrodes are lightweight and can be easily integrated into the prosthesis.
However, they can elicit unnatural sensations often considered unpleasant, high electrical voltage is required, and the presence of electrical fields can interfere with the myoelectric measurements necessary for prosthesis control \cite{see2022touch,bensmaia2020restoration}.

On the other hand, vibrotactile stimulation has emerged as one of the most extensively investigated techniques for its ease in tactile signal modulation and for vibrotactile actuators' compact size, affordability, and availability~\cite{choi2012vibrotactile,thomas2019comparison}.
Various methods have been suggested for converting a prosthetic hand's contact cue with an object and grip force into vibrotactile actuation, such as in \cite{raveh2018myoelectric,markovic2018clinical}. 
However, despite the positive outcomes in terms of performance time in manual dexterity tests, grip, and manipulation accuracy, force vibrotactile feedback systems are not always integrated within the prosthetic device. 
Indeed, the vibrotactile actuators are usually placed with elastic bands on the arm or wrist of able-body subjects or on prosthetic users' contralateral arm and residual limb \cite{raveh2018myoelectric,markovic2018clinical}.
When using a prosthetic device in daily living, having a separate feedback system, such as an additional device, can be uncomfortable and inconvenient due to lack of integration.

According to Kim et al. \cite{kim2012haptic} an optimal haptic feedback system should ensure the somatotopic matching (SM) and the modality matching (MM) paradigms. 
The somatotopic-matched feedback creates a natural signal comparable to the original sensation for the same body part. 
Antfolk et al. \cite{antfolk2012sensory} have demonstrated that sensory perception due to contact stimuli at the stump level can be processed by the central nervous system similar to that elicited by stimuli at the finger level. 
On the other hand, the modality-matched feedback provides a stimulus similar to the original sensation, i.e. pressure stimuli as a force cue. 
Vibrotactile stimuli are naturally associated with high-frequency information arising from surface contact, such as texture, roughness and first-contact cues \cite{visell}.
Therefore, under the MM paradigm, the use of vibrotactile stimulation should convey acceleration-mediated contact cues related to the characteristics of objects being touched and facilitate their manipulation \cite{actionsomatosensory}. 
Nevertheless, methods to reproduce cutaneous cues of the surfaces as texture for prosthetic users are less investigated than solutions to communicate contact and grip force of prosthetic limbs \cite{balamurugan2019texture}. 
Furthermore, to the best of the authors’ knowledge, commercial embedded vibrotactile prosthetic feedback solutions for texture and contact artificial perception are still not currently available.

Methods to deliver texture and contact cues include the Rehand II \cite{sakuma2020electric}. 
The device is endowed with a skin vibration sensor and three vibrotactile actuators.
The system undergoes testing with eight able-bodied adults, yielding favourable outcomes in relation to contact and operation recognition tasks.
One of the very few integrated and commercialized vibrotactile feedback is presented in \cite{akhtar2020touch}.
The PSYONIC Ability Hand is endowed with a single vibrator motor.
Pressure sensors are placed on all five digits and mapped into the motor.
While the PSYONIC Ability Hand is an advanced and sensorized prosthesis that conforms to the somatotopic paradigm, matching pressure cues with motor vibration could potentially increase the cognitive load needed to interpret haptic feedback.

In our pursuit to advance tactile feedback technology with an intuitive, wearable, and embedded solution, we present a novel Vibro-Inertial Bionic Enhancement System (VIBES) (Fig.\ref{fig:system}) in this work integrated with a SoftHand Pro (SHP) \cite{godfrey2018softhand} prosthetic hand. 
In our previous work \cite{paperfani}, we tested a MM vibrotactile system that recorded accelerations from the fingers of an artificial hand and reproduced them on a prosthetic user’s skin through voice-coil actuators. 
In this work, the feedback system comprises two Inertial Measurement Units (IMUs) on the prosthetic thumb and index fingernails as sensors and two integrated vibrotactile actuators, efficiently transmitting texture and contact cues to specific stump sites to achieve both MM and SM. 
The presence of intrinsic somatosensory feedback generated by artificial body parts has been shown to be crucial for accurate motor commands \cite{amoruso2022intrinsic}. 
Thus, incorporating vibrotactile actuators in direct contact with the stump not only complies with the MM paradigm but also addresses the absence of intrinsic somatosensory feedback caused by the damping elements present in the SHP.
\section{System Design}\label{sec:1}
The system is integrated inside the inner socket, with two vibrotactile actuators.
Fig.\ref{fig:system} presents an overview of the VIBES.
As in our previous work \cite{barontini2021wearable}, the mechanical structure consists of three components: the structural frame, the mechanical actuation, and the feedback interface. 
The SHP, the socket, and the inner socket containing surface electromyographic (sEMG) sensors make up the structural frame (parts {1, 2, 3, 5} in Fig.~\ref{fig:system}).
The mechanical part comprises a battery pack (6), an electronic board (7) and two IMUs (8). 
Two planar vibrotactile actuators (4) are integrated into the inner socket and part of the feedback interface.
\subsection{Vibrotactile Actuators}
The vibrotactile actuators used in this study are Haptuator Planar (HP) actuators by TactileLabs\footnote{Haptuator Planar by TactileLabs, [Online], Available: \url{http://tactilelabs.com/products/haptics/haptuator-planar/}} (See Fig.~\ref{fig:control} on the right).
The HP is made of a coil enclosing a permanent magnet.
It has a 50-500 Hz high bandwidth which is within the range of tactile sensitivity.
The transmission of stimuli is normal to the skin to reduce the size of the system with respect to tangential stimuli transmission.
Its lightweight (1.8 gr), compact shape (12x12x\SI{6}{mm}), and soft surface designed to be directly in contact with the skin make the HP suitable for integration into prosthetic devices.
\subsection{The control strategy} \label{sec:control}
\begin{figure}
      \centering
      \includegraphics[width=\columnwidth]{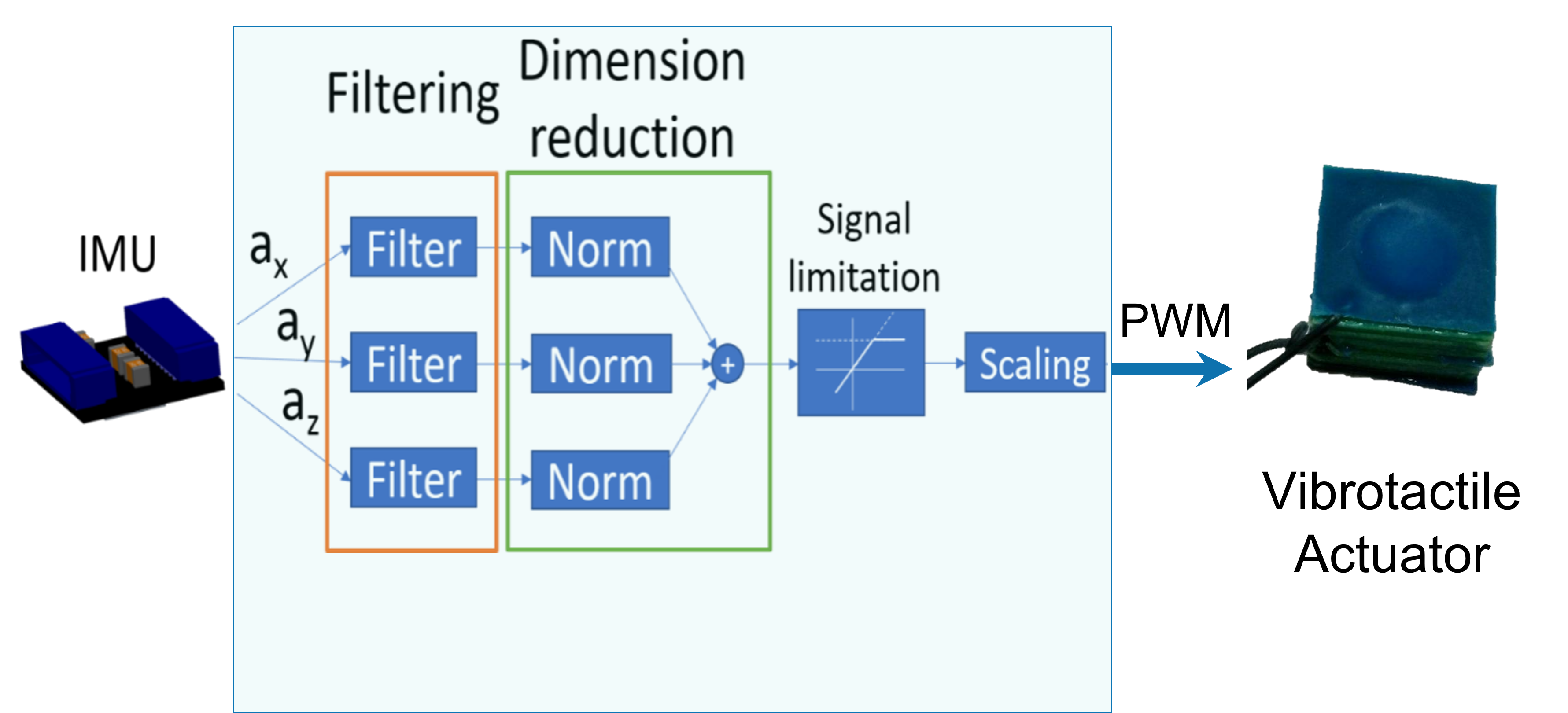}
      \caption{Control strategy diagram of the vibrotactile system. The acceleration signals acquired from the IMU~(left) undergo a signal elaboration process (filtering, dimensional reduction, and other minor elaboration steps). The processed signals are encoded into a PWM signal and transmitted to the actuator~(right).}
      \label{fig:control}
      \vspace{-0.5cm}
\end{figure} 
The control strategy of the VIBES is shown in Fig \ref{fig:control}.
The acceleration signals $a = (a_x,a_y,a_z)$ recorded by each IMU are filtered, dimensional reduced and limited as in \cite{paperfani}. 
The filtering stage aims at removing unwanted artefacts caused by accelerometer readings during free-hand motion and vibrations from the prosthesis actuation system.
Then, a scaling factor is applied to map the acceleration signals to a value of PWM to activate the actuator, considering the actuators' current limits. 
To achieve the somatotopic paradigm, the index IMU signal is mapped to the left actuator and the thumb IMU signal to the right actuator.
The control works in real-time, and electronic boards \cite{NMMI} are used to record signals and drive the actuators. 
For further details about the control algorithm, see our previous work in \cite{paperfani}.
\section{System Characterization}\label{sec:2}
To test the effectiveness of the actuators in conveying reliable vibrotactile cues, we perform a physical and psychophysical characterization of the system.
\subsection{Physical characterization}\label{sec:2a}
\begin{figure}
      \centering
      \includegraphics[width=\columnwidth]{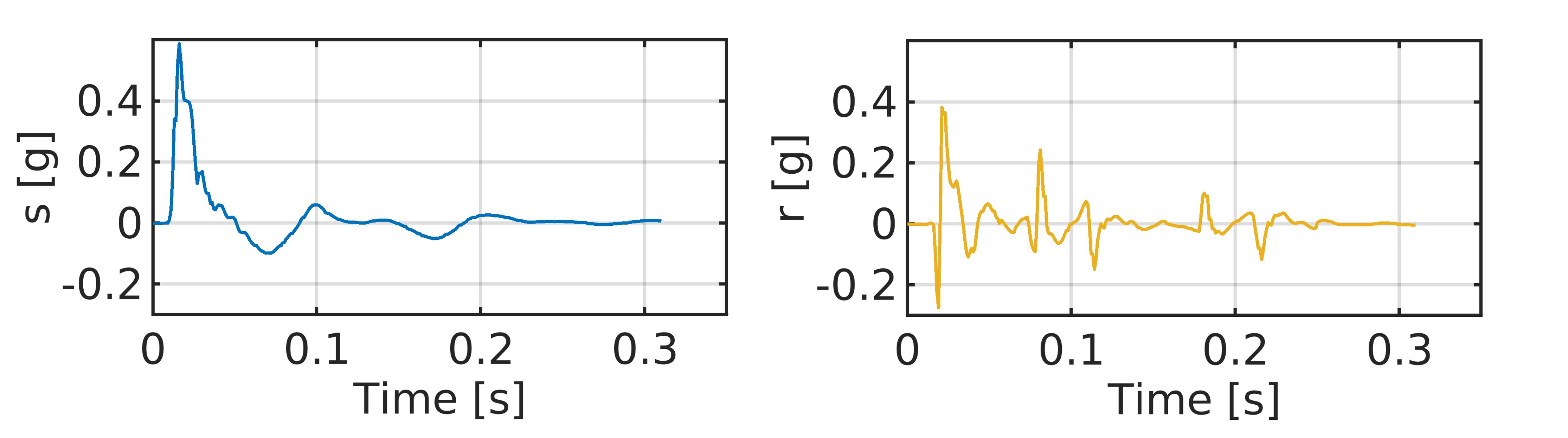}
      \caption{Comparison between an acceleration signal $s$ (left) and its rendering by the HP actuator $r$ (right).}
      \label{fig:signals}
      \vspace{-0.3cm}
\end{figure}

The HP performance in rendering tactile input signals is tested. 
During tactile exploration, we save acceleration signals from an IMU on the robotic hand.
Then, we measure the HP rendering by placing another IMU directly on the actuator.
Fig.~\ref{fig:signals} reports the comparison between an acceleration signal $s$ and its rendering $r$.
In this phase, no filtering techniques or signal reduction are applied.
Based on the results obtained, which show the system's attempt to replicate the trend of the reference input signal, we set the control algorithm in Section \ref{sec:control}.
\subsection{Psychophysical characterization}\label{sec:2b}
The experiment evaluates the capacity of human participants to discriminate different levels of roughness rendered by the actuators, corresponding to different sandpaper signals, in terms of Just Noticeable Difference (JND).
JND refers to the minimum change in a stimulus required to produce a detectable difference in sensation~\cite{jones2012application}. 
We use the method of constant stimuli \cite{jones2012application}, a common psychophysical technique, to calculate JND by presenting two stimuli and asking the participant to judge which one is greater.
The experiment includes a \textit{recording session} and a \textit{user session}. In the \textit{recording session}, the external tactile stimuli are recorded from the index finger of the SoftHand Pro and then saved.
The index finger is chosen as the most commonly used finger for tactile exploration. 
In this stage, no filtering techniques are used. 
In the \textit{users' session}, the tactile stimuli are reproduced on the users' fingers with the HP.

All the experimental procedures are approved by the Committee on Bioethics of the University of Pisa-Review N. 30/2020.
\subsubsection{Participants}
Fifteen able-bodied right-handed participants (7 females and 8 males, age mean$\pm$SD: 26,6$\pm$2,05) are enrolled and asked to give their informed consent to participate in the experiments. 
Participants have no disorder that could affect the experimental outcome.
\subsubsection{Recording session}
Five discrete stimuli are recorded in the \textit{recording session}.
The signal comes from sliding  the SoftHand Pro index finger on different sandpapers.
The set of sandpapers comes from the same manufacturer, RS\footnote{RS, [Online], Available: \url{https://it.rs-online.com/web/}}, and grit numbers and particle sizes shown are according to the Federation of European Producers of Abrasive Products (FEPA) P-grading system.
\begin{table}
\small
\MFUnocap{for}%
\MFUnocap{the}%
\MFUnocap{of}%
\MFUnocap{and}%
\MFUnocap{in}%
\MFUnocap{from}%
\caption{\capitalisewords{The five sandpapers selected: FEPA P-GRADE and average grit size.}}
\label{tab:grit}
\begin{center}
\begin{tabular}{l|l|l}
\hline
       &  FEPA P-GRADE                     & Average grit size ($\mu$m) \\ \hline
       Smooth & P1000 & 18  \\ 
        & P220 & 65  \\ 
        Reference & P120 & 127  \\ 
         & P80 & 195  \\ 
         Rough & P60 & 264  \\ \hline
\end{tabular}
\vspace{-0.6cm}
\end{center}
\end{table}
A preliminary study is done to determine the stimuli range.
The smoothest and the roughest sandpapers detectable by the actuator are identified.
The average grit sizes vary from 18 $\mu$m (grit number P1000) to 264 $\mu$m (grit number P60). 
To select the five stimuli, following \cite{libouton2010tactile} procedure, we decide to use as reference stimulus P120 (average particle size of 127 $\mu$m), about halfway between the smoothest and the roughest stimuli.
The other two stimuli are chosen to keep the stimuli equally spaced (in terms of micrometres), considering the existing types of sandpapers.
Furthermore, according to the method of constant stimuli \cite{jones2012application}, the extreme stimuli are chosen to be easily distinguishable from the reference stimulus, while the other stimuli are intentionally made challenging to differentiate.
The selected sandpaper average grit sizes and grit numbers are shown in Table \ref{tab:grit}.
Hence, five different sandpapers of 28x23 cm are used to record the signals.

\begin{figure}
      \centering
      \includegraphics[width=\columnwidth]{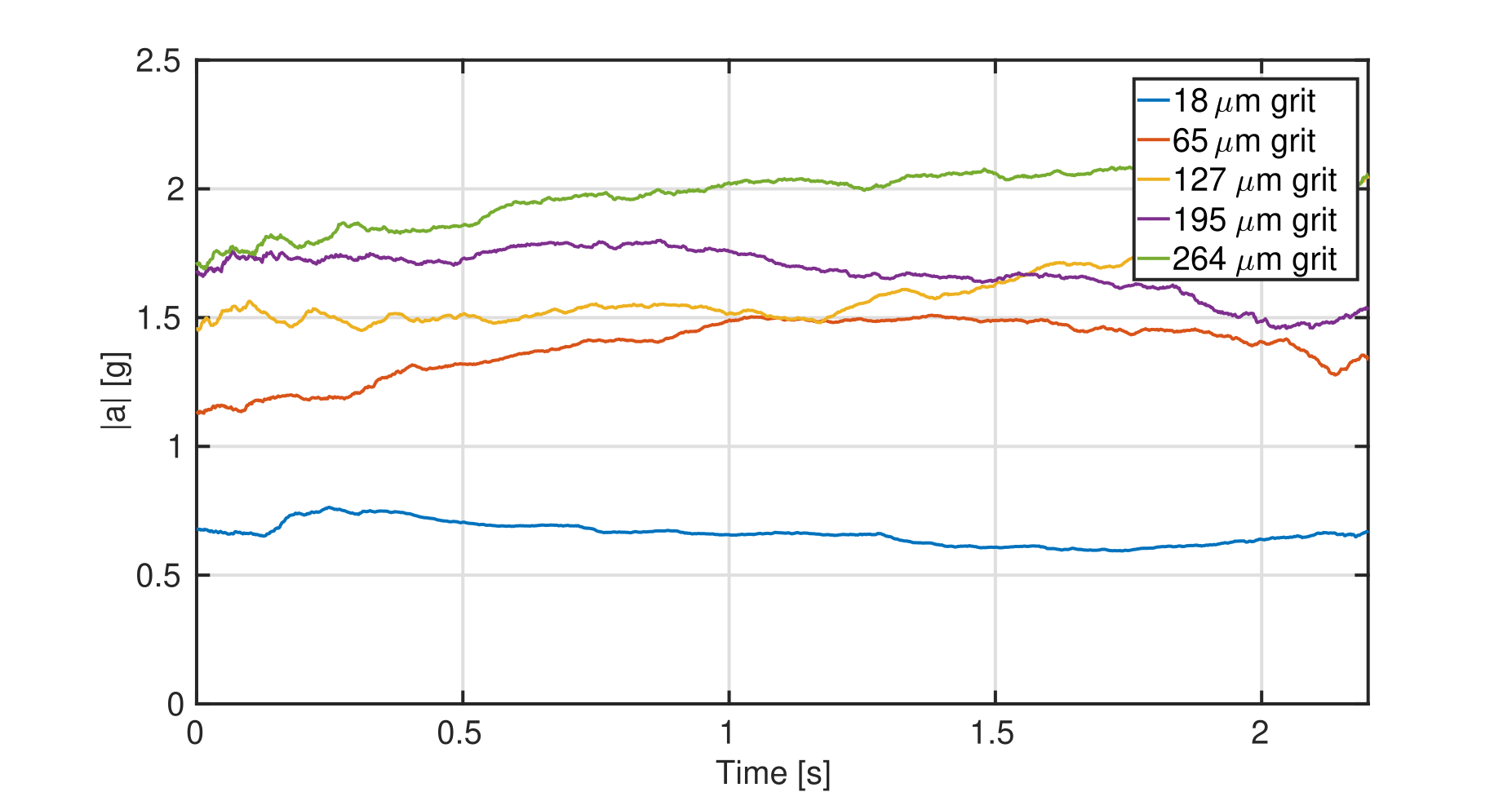}
      \caption{Sandpaper acceleration signals: acceleration module $|$a$|$ recorded for each sandpaper type.}
      \label{fig:sp_signals}
      \vspace{-0.6cm}
\end{figure}

A custom C++ software is developed to register, cut, and save the accelerations coming from the IMU on the index fingertip. 
The three components of the acceleration $ a = ( a_x, a_y, a_z )$ are recorded from the IMU.
The experimenter tries to keep velocity and pushing force as constant as possible by keeping fixed the time window to complete the recordings and taking the SHP at the same height above the paper.
As reported in \cite{boundy2017speed}, passive texture presentation at varying scanning speeds had minimal effect on perception.
Moreover, the perception of texture is also invariant to changes in contact force \cite{SAAL201899}.
Thus, velocity magnitude and pushing force should not influence the experiment results.
Finally, two signals per sandpaper type are saved.
Fig. \ref{fig:sp_signals} shows the module of five different sandpaper signals recorded. 
A moving average filter is applied to improve the reading of the graph and facilitate the comparison.
Each acceleration signal is mapped to a value of PWM to activate the HP, following the control strategy in Section \ref{sec:control}. 
\subsubsection{User session}
In the \textit{user session}, participants are comfortably seated on an office chair, with their arm (elbow and forearm) placed on a desk. 
A custom 3D printed case is designed to enable the participants to use the HP comfortably.
Participants are asked to keep their right index finger on the HP.
The participants are isolated by wearing goggles with obscured lenses and headphones playing white noise.
The experimenter verifies that the index finger lays on the case without putting pressure on it and without changing the finger position.
A custom C++ software is developed to automatically manage the experiment following the requirements of the psychophysical method chosen.

The test involves presenting a reference and a comparison stimulus to the participant, with an audible signal announcing each pair. 
The reference stimulus is always P120, and the participant must indicate which stimulus felt rougher. 
There is no familiarization phase, and stimuli are not repeated.
The comparison signal is randomly selected from the five pre-registered signals and presented to the subject in a pseudo-randomized order.
Moreover, both the reference and the comparison signals are randomly chosen from the two stimuli saved for each sandpaper type.
Each experiment consisted of 100 trials, 20 trials per stimulus level (corresponding to the five sandpaper grit).
\subsubsection{Data Analysis}
By means of a Generalized Linear Mixed Model (GLMM), we test the variability of the perceived roughness of the 15 subjects.
GLMM account for the effect of the experimental variables of interest and the heterogeneity between clusters through fixed- and random-effect parameters, respectively \cite{10.1167/12.11.26}. We used the following model:
\begin{equation}
\begin{split}
\phi^{-1}&[P(Y_j=1)]= \beta_0 + \beta_1*x_j,
\label{eq:glmm}
\end{split}
\end{equation}
where $\phi^{-1}[\cdot]$ is the probit link function, which is due to dependent dichotomous variable, $P(Y_j=1)$ is the probability of perceiving the comparison stimulus as rougher than the reference in trial j, $\beta_0$ and $\beta_1$ are the fixed effect parameters, i.e. the intercept and the slope of the linear function (linear predictor), which are the same for all the subjects. The explanatory variable is  \textit{$x_j$}, i.e. the sandpaper stimuli.
Next, we estimated the JND and the Point of Subjective Equality (PSE), which is the stimulus value yielding a response probability of 0.5, with the related 95\% confidence intervals (CIs), using the bootstrap method \cite{10.1167/12.11.26}.
JND and PSE were computed as in \cite{10.1167/12.11.26}:
\begin{equation}
\begin{split}
JND =\frac{1}{\beta_1},
\label{eq:JND}
\end{split}
\end{equation}
\begin{equation}
\begin{split}
PSE = -\frac{\beta_0}{\beta_1}.
\label{eq:PSE}
\end{split}
\end{equation}
\begin{figure}
      \centering
      \includegraphics[width=0.85\columnwidth]{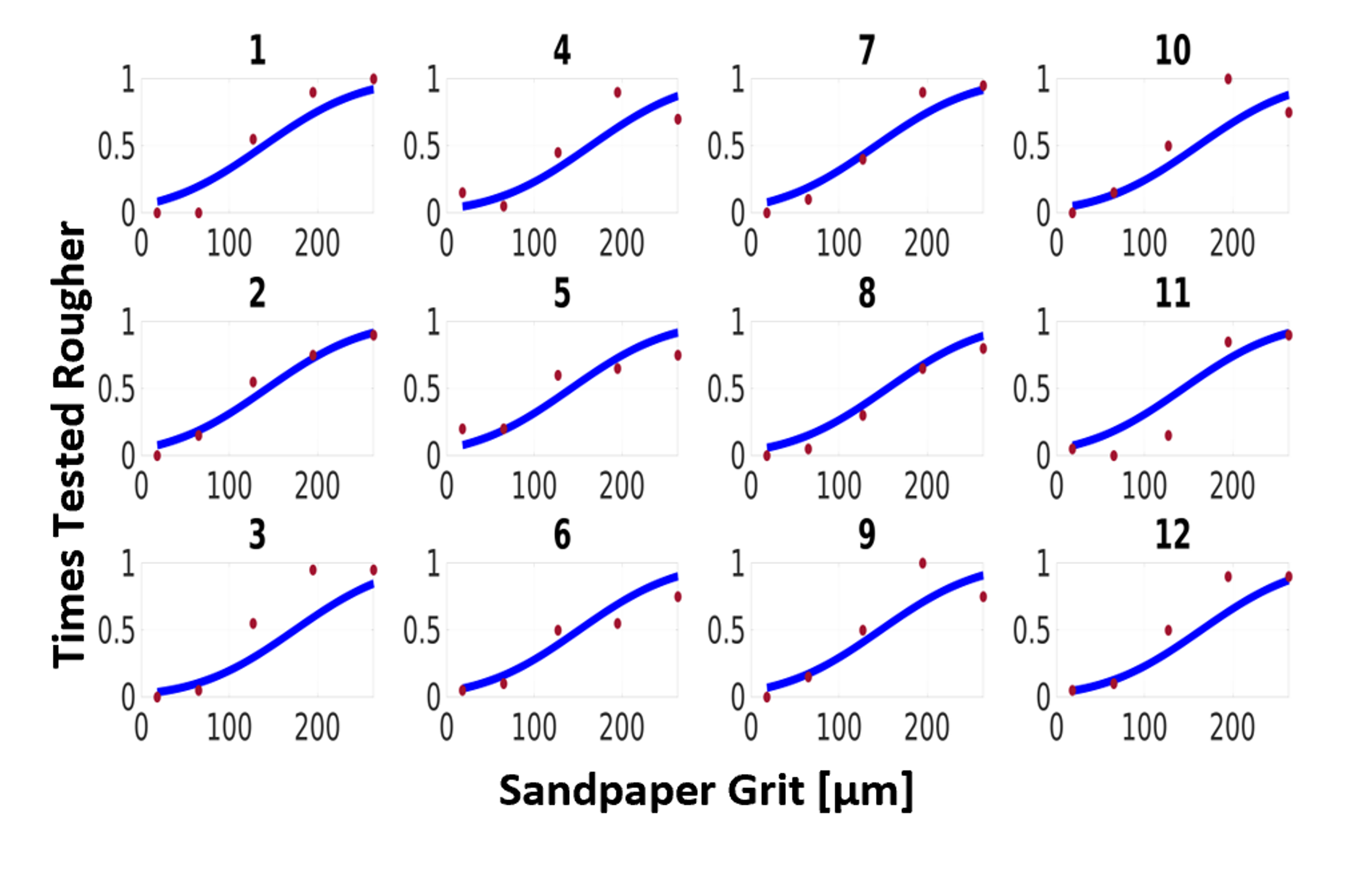}
      \caption{GLMM fit for the 12 subjects. Raw data and model predictions for each participant labeled as 1 to 12.}
      \label{fig:jndhs}
\end{figure}
\begin{figure}
      \centering
      \includegraphics[width=0.9\columnwidth]{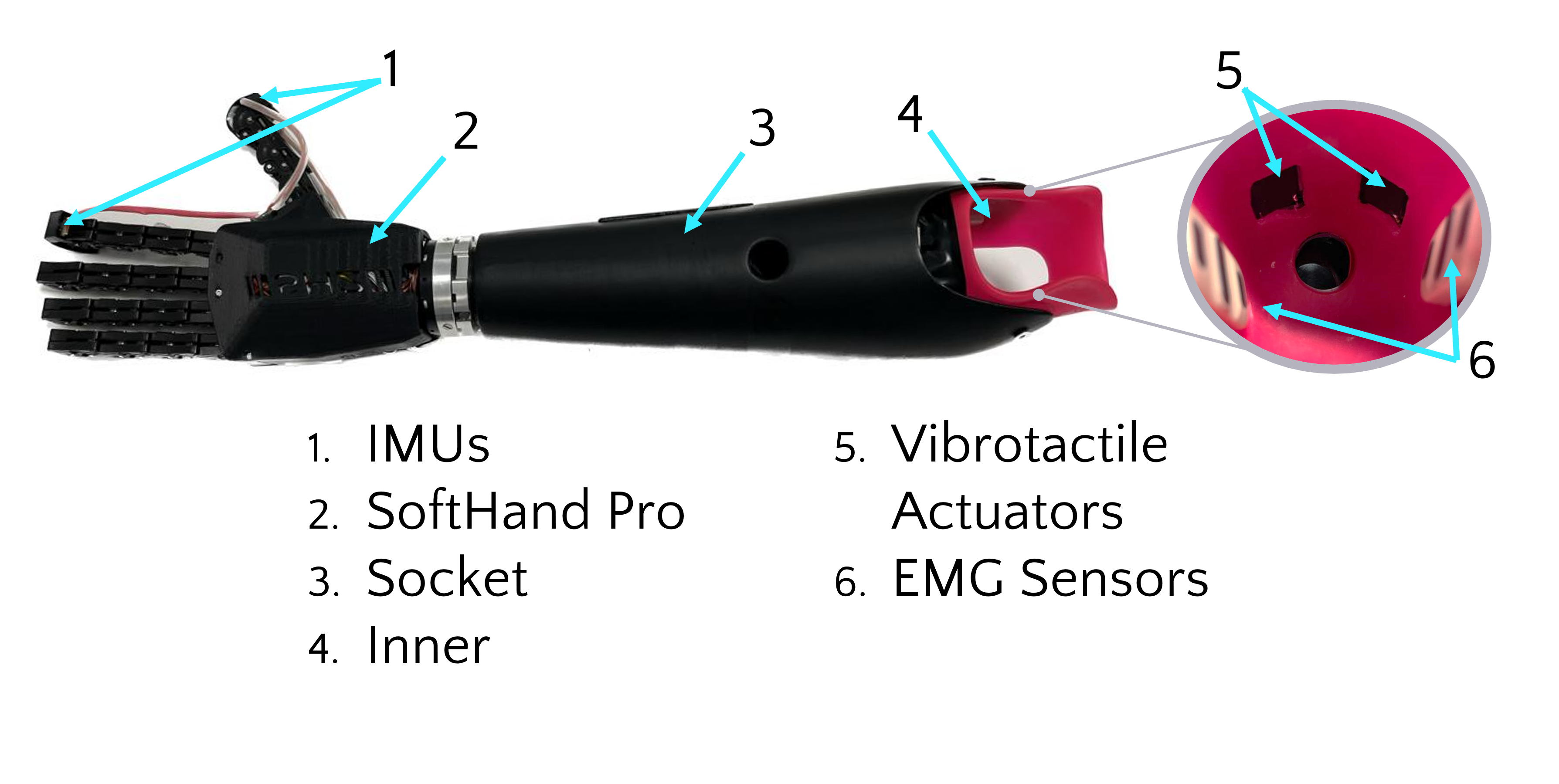}
      \caption{The VIBES integrated inside the prosthesis. A detailed view (right) shows the inner part of the socket with vibrotactile actuators and EMG sensors.}
      \label{fig:vibesreal}
      \vspace{-0.6cm}
\end{figure} 
\subsection{Results}
The GLMM fitting to the data is illustrated
in Fig. \ref{fig:jndhs}.
We evaluated the capacity of participants to discriminate roughness stimuli rendered by the HP actuator.
We model 12 subjects' data as in (\ref{eq:glmm}). 
Three outliers were identified, exhibiting flat curves and indicating a low sensitivity to the texture (i.e. the subjects did not perceive significant distinctions among the five stimuli).
The JND was equal to 87.30 $\mu$m (95\% CIs:  79.69 - 96.5251 $\mu$m).
The PSE was equal to 151.96 $\mu$m (95\% CIs: 122.77 - 187.32 $\mu$m).
\section{Pilot Experiments}\label{sec:3}
We tested the VIBES with one prosthetic user (age 43, female) affected by agenesis of the left forearm. 
The subject does not suffer from any cognitive impairment that could have interfered with her ability to follow the instructions of the study.
The participant usually wears a cosmetic prosthesis, although she has experience with myoelectric ones.
We integrated the VIBES inside an inner socket.
Due to spatial constraints within the inner socket resulting from the presence of EMG sensors, we positioned both HP actuators on the hairy skin. This arrangement was chosen to effectively address the limited space available and ensure the appropriate placement of the actuators within the designated configuration.
Fig. \ref{fig:vibesreal} shows the VIBES inside the inner socket.
Experiments aim to assess the efficacy of the feedback provided by the VIBES device with the SoftHand Pro in isolated conditions to reflect low-light situations and allow users to focus on the feedback~\cite{raveh2018myoelectric}.
The tests assess roughness discrimination and identification abilities by transmitting first-contact cues to the subject.
For the SHP, the same control strategy as in \cite{barontini2021wearable} is used.
At the end of the experiments, the subject completes the System Usability Scale (SUS) to evaluate the system \cite{Lewis2018TheSU}.

All the experimental procedures are approved by the Committee on Bioethics of the University of Pisa-Review N. 30/2020.
\subsection{Psychophysical Characterization Experiment}
A psychophysical characterization of a prosthetic user is performed to assess whether the subject’s perception is similar to that of able-bodied individuals.
The experimental protocol for the prosthetic user is the same as the one used for the able-bodied participants in Section \ref{sec:2}.
Thus, only the left HP is activated.
\subsection{Active Texture Identification Experiment}
\begin{figure}
      \centering
      \includegraphics[width=\columnwidth]{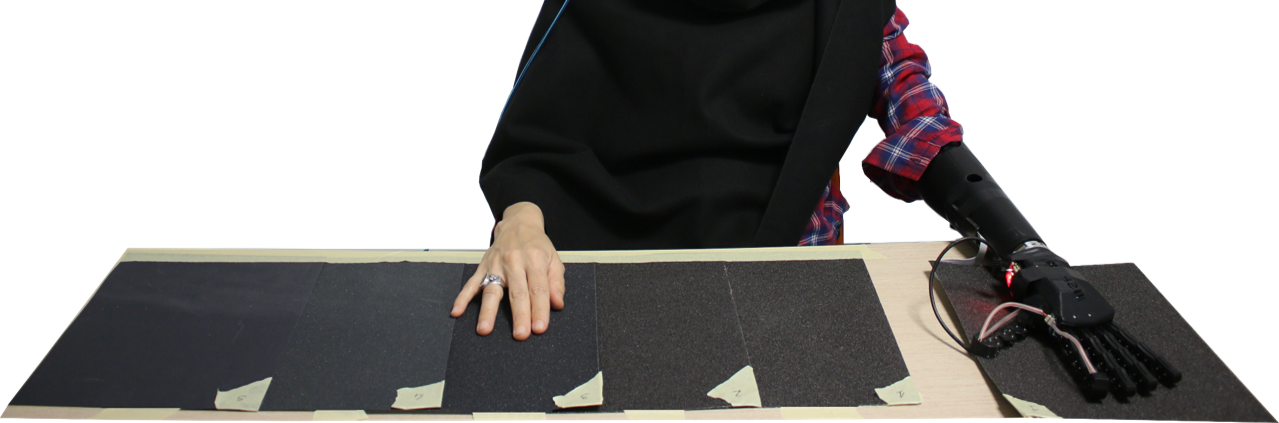}
      \caption{The prosthetic user during the Active Texture Identification Experiment.}
      \label{fig:expsetup}
       \vspace{-0.3cm}
\end{figure}

For active texture identification, the subject is seated on a chair in front of a table.
First, during 10 minutes familiarization phase, the participant is allowed to freely explore a P150 sandpaper (different from the one used in the subsequent experiment) fixed on the table while the VIBES is turned on.
No isolation is provided in this phase.
Then, during the experiment, the subject is isolated with white noise and obscured lenses. 
Fig. \ref{fig:expsetup} shows the setup of the experiment. 
The same sandpapers used in Section \ref{sec:2} are used in this task.
The set of five sandpapers in descending roughness order, numbered from 1 (P60) to 5 (P1000), is placed on the right side of the table, which the subject explores using the right hand. 
On each trial, one sandpaper is presented to the subject under the left prosthetic hand for exploration using the prosthetic device. 
The participant is asked to identify the matching sandpaper from the five options on the right side. 
The task is performed in two modalities, with and without the VIBES feedback, and each sandpaper is randomly presented five times for a total of 25 identifications. 
Confusion matrices and accuracy metrics are used to analyze subject performance\cite{grandini2020metrics}. 
\subsection{Results}
\begin{figure}
\includegraphics[width=\columnwidth]{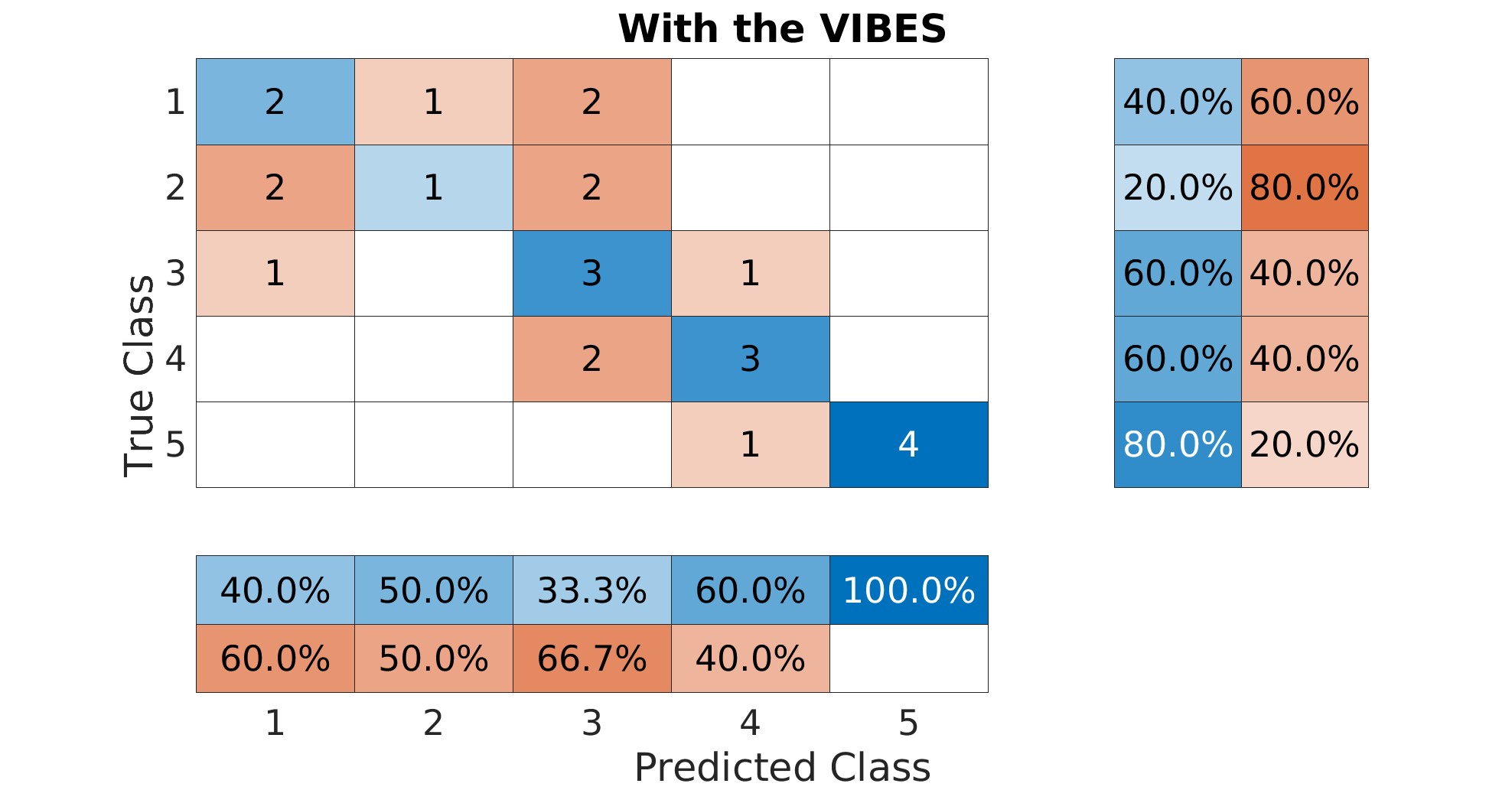}\quad
\includegraphics[width=\columnwidth]{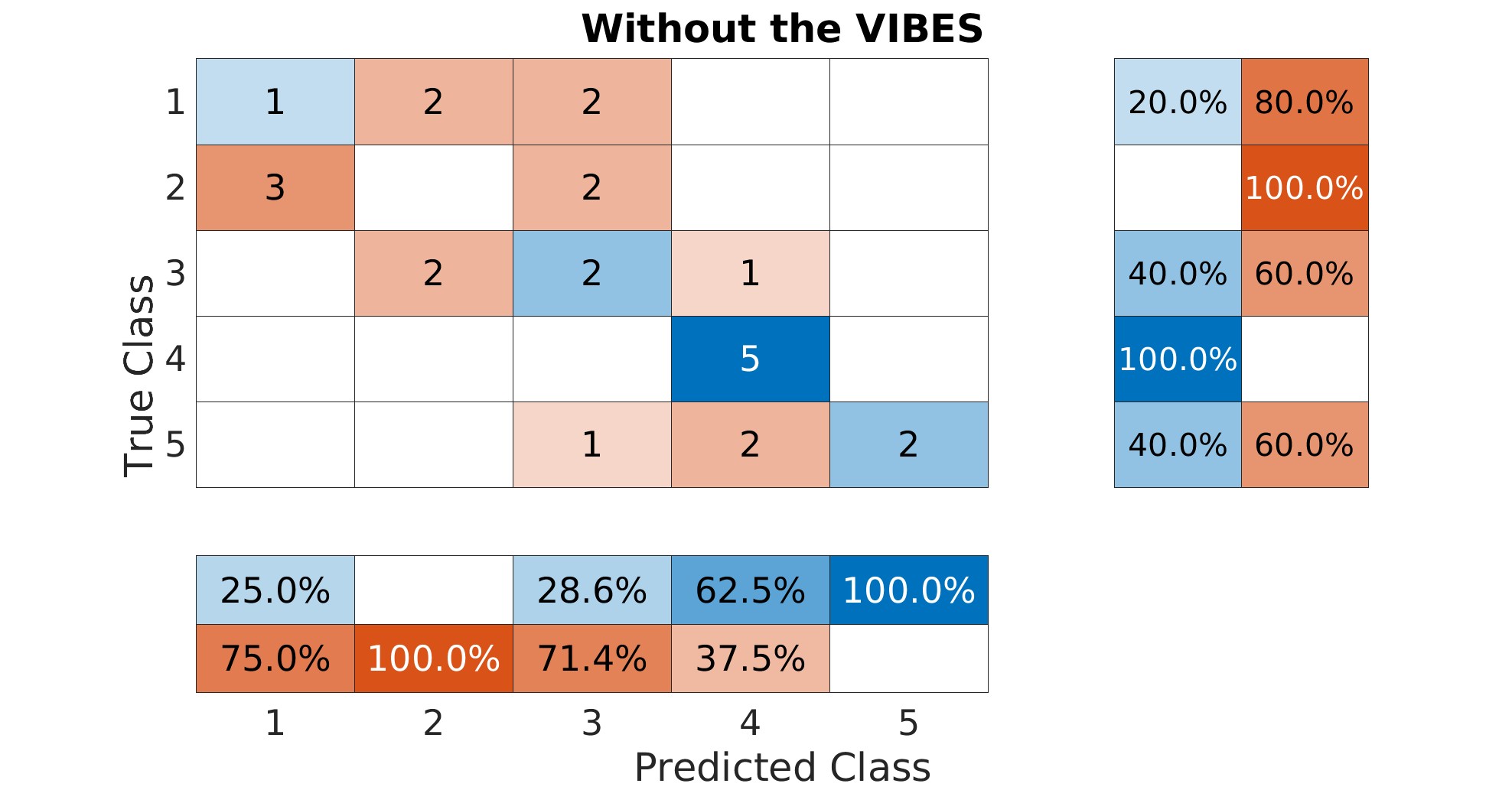}\quad
\caption{Active Identification Experiment: confusion matrices with and without the VIBES. Stimuli in descending roughness order, from 1 (P60) to 5 (P100) (see Table \ref{tab:grit}). The row and column summary displays the percentage of correctly classified and incorrectly classified observations for each true or predicted class.}
\label{fig:ati}
\vspace{-0.5cm}
\end{figure}

Since only one subject was involved, no statistical analysis was performed for the psychophysical characterization experiment.
The accuracy results are shown in Table \ref{tab:psy}; from a probit fit on the data, the JND was 44.07 $\mu$m.
\begin{table}[]
\small
\MFUnocap{for}%
\MFUnocap{the}%
\MFUnocap{of}%
\MFUnocap{and}%
\MFUnocap{in}%
\MFUnocap{from}%
\MFUnocap{with}%
\centering
\caption{\capitalisewords{Results of the comparison between the pair stimuli with the percentage of success.}}
\label{tab:psy}{%
\begin{tabular}{clc}
\textbf{Comparison} & \textbf{I \textgreater Ref} & \textbf{\% Success} \\ \hline
1-3                 & 0                           & 100\%               \\
2-3                 & 10                          & 50\%                \\
3-3                 & 7                           & 65\%                \\
4-3                 & 19                          & 95\%                \\
5-3                 & 20                          & 100\%              
\end{tabular}%
}
\vspace{-0.6cm}
\end{table}
The results of the Active Texture Identification experiments are shown in Fig.\ref{fig:ati}.
The confusion matrices show a major improvement in sandpaper identification when the subject wears the feedback device with respect to without feedback.
The overall accuracy with the VIBES was 52\%, and without the VIBES was 40\%.
Furthermore, the SUS questionnaire resulted in a positive score of 77.5 (average SUS score of 68 at the 50$^{th}$ percentile).

\section{Discussion}\label{sec:4}
The Psychophysical Characterization of the VIBES revealed consistent behaviours among able-bodied subjects when discriminating roughness stimuli.
JND was equal to 87.30 $\mu$m, in line with the results in \cite{libouton2010tactile} regarding human index fingertip exploration. 
Indeed, the tactile roughness discrimination threshold at the index fingertips in humans with active exploration is 15$\pm$8.5 $\mu$m (smoothed sandpapers) and 44$\pm$32.5 $\mu$m (roughest sandpapers)\cite{libouton2010tactile}.
Considering that our experiments involved passive exploration of texture through vibrotactile actuators rendering, our results are in accordance with these prior findings.
Furthermore, the Psychophysical Characterization Experiment with a prosthetic user yielded positive results, with a Just Noticeable Difference of 44.07 $\mu$m.
Thus, the HP actuator is able to enable users to perceive and discern roughness with high precision.

The Active Texture Identification Experiment with a prosthetic user yielded favourable outcomes.
The participant could better identify the sandpaper presented with the VIBES feedback than without it.
Indeed, the accuracy with the vibrotactile feedback was 52\% with respect to 40\% without the feedback. 
Although the study results are promising, achieving 100\% accuracy was not possible. 
This can be attributed to the difference in skin sensitivity between the stump and fingers, as there are more receptors on the fingers, which could potentially decrease the subject's discrimination performance.
To support this hypothesis, an alternative modality of the experiment could have involved exploring various textures using the residual limb without the prosthetic device. 
However, we have chosen to recreate a realistic everyday scenario in which the subject, accustomed to wearing a prosthesis, can effectively explore the external environment with the prosthesis.
It is also important to note that even able-bodied individuals may not have 100\% accuracy in discrimination between similar roughness using their own fingers~\cite{motamedi2016use}. 

Additionally, the subject's ability to identify the matching sandpaper without feedback was higher than chance, which could be due to her expertise in wearing a myoelectric prosthesis and intrinsic and extrinsic feedback propagation in recognizing different textures.
Despite the presence of damping elements in the SoftHand Pro, it is possible that the transmission of vibrations through the socket could have affected the subject's perception.
More tests are required to delve further into this aspect.

Based on the subject's comments, the VIBES provided an intuitive experience, and this was further supported by a high score on the SUS questionnaire, which is a promising indication of the VIBES usability and user satisfaction.

It is worth noticing that the results obtained from our study are preliminary, and we acknowledge the need for further research. 
We aim to conduct a more in-depth physical characterization of the actuators, focusing on analyzing the correlation between the input signal and the resulting signal produced by the actuators.
This will provide a deeper understanding of their properties and capabilities.
Furthermore, future investigations will comprehensively study how the actuator's position on the stump (e.g. on the hairy or on the glabrous skin) influences how individuals perceive sensory stimuli.
Moreover, we will investigate the effect of contact and texture cues even on manual dexterity tasks and with a large sample size to conduct statistical analyses.

It is worth mentioning that the VIBES completes a SoftHand Pro framework, providing the unique advantage of dual haptic feedback options.
Indeed, in our previous research \cite{barontini2021wearable}, we introduced a soft pneumatic integrated system, the WISH device, which effectively conveyed contact and grip force cues to prosthetic users by applying pressure stimuli on their stumps.
The WISH device and the VIBES transform the SHP into a highly adaptable prosthesis, empowering the user to select their preferred type of feedback.
The comprehensive results derived from the experiments with both able-bodied and limb-loss participants show that the VIBES is a promising solution for transmitting and restoring tactile feedback.
\section{Conclusion}\label{sec:5}
This study introduces the VIBES device, a vibrotactile integrated feedback system to transmit texture and contact cues to prosthetic users. 
To evaluate the device's effectiveness, we characterized the system with 15 able-bodied subjects with positive results in terms of texture discrimination performance. 
From the usability experiments with a prosthetic user, it can be concluded that the VIBES can effectively transmit tactile feedback.
The subject was able to use the feedback to improve the success rate in almost all proposed trials.
Nonetheless, further research is needed to fully evaluate the device's effectiveness and potential.
The positive feedback from the participants, combined with the results of this study, motivate us to conduct further investigations.


\section*{ACKNOWLEDGMENT}
The authors would like to thank Marina Gnocco, Mattia Poggiani and Manuel Barbarossa for their valuable support in the experiments. 

\bibliographystyle{IEEEtran}
\bibliography{mybibfile}
\tiny
\end{document}